\documentclass{article}

\usepackage[final]{nips_2017}

\usepackage[utf8]{inputenc} 
\usepackage[T1]{fontenc}    
\usepackage{hyperref}       
\usepackage{url}            
\usepackage{booktabs}       
\usepackage{amsfonts}       
\usepackage{nicefrac}       
\usepackage{microtype}      
\usepackage[pdftex]{graphicx}
\usepackage{amsmath,amssymb}
\usepackage[boxed]{algorithm2e}
\usepackage{mathtools}
\usepackage{enumerate}

\DeclareMathOperator*{\argmax}{arg\,max}

\title{Efficient exploration with \\ Double Uncertain Value Networks}

\author{
  Thomas M. Moerland, Joost Broekens and Catholijn M. Jonker \\
  Department of Computer Science\\
  Delft University of Technology, The Netherlands \\
  \texttt{\{T.M.Moerland,D.J.Broekens,C.M.Jonker\}@tudelft.nl} \\
}

\begin{document}

\maketitle

\begin{abstract}
This paper studies directed exploration for reinforcement learning agents by tracking uncertainty about the value of each available action. We identify two sources of uncertainty that are relevant for exploration. The first originates from limited data ({\it parametric uncertainty}), while the second originates from the distribution of the returns ({\it return uncertainty}). We identify methods to learn these distributions with deep neural networks, where we estimate parametric uncertainty with Bayesian drop-out, while return uncertainty is propagated through the Bellman equation as a Gaussian distribution. Then, we identify that both can be jointly estimated in one network, which we call the Double Uncertain Value Network. The policy is directly derived from the learned distributions based on Thompson sampling. Experimental results show that both types of uncertainty may vastly improve learning in domains with a strong exploration challenge.
\end{abstract}

\section{Introduction}
Reinforcement learning (RL) is the dominant class of algorithms to learn sequential decision-making from data. In RL we start with zero prior knowledge and need to actively collect our own data. Therefore, we should not settle on a policy too early, instead of trying out actions we have not properly explored yet. However, we neither want to continue exploring sub-optimal actions, when we already know what is best. This challenge is known as the exploration/exploitation trade-off.

Most state-of-the-art reinforcement learning implementations use {\it undirected} forms of exploration, such as $\epsilon$-greedy or Boltzmann exploration. These methods act on {\it point estimates} of the mean action-value, usually applying some random perturbation to avoid only selecting the currently optimal action. However, these undirected methods are known to be highly inefficient \citep{osband2014generalization}. By only tracking point estimates of the mean state-action value, these algorithms lack the information to, for example, discriminate between an action that has never been tried before (and requires exploration) and an action that has been tried extensively and deemed sub-optimal (and can be avoided).

A natural solution to this problem originates from tracking uncertainties/distributions. The intuition is that with limited data and large uncertainty there is reason to explore, while narrow distributions naturally transfer to exploitation (see Appendix \ref{illustration} for a detailed illustration). For this work we identify two types of uncertainties/distributions that are interesting for exploration: 
\begin{itemize}
\item {\it Parametric uncertainty}: This is the classical statistical uncertainty which is a function of the number of available data points. The cardinal example is the posterior distribution of the mean (action-value). 
\item {\it Return uncertainty}: This is the distribution over returns from a state-action pair given the policy. For this work we focus on deterministic domains, which makes the return distribution entirely induced by the (exploratory) stochastic policy. 
\end{itemize} 

We argue that - for deterministic environments - we can explore by acting probabilistically optimal with respect to {\it both} distributions (see Section \ref{valuefunctions}). We identify neural network methods to estimate each of them separately, and subsequently show that both can be combined in one network, which we call the Double Uncertain Value Network (DUVN). To the best of our knowledge, we are the first to 1) distinguish between uncertainty due to limited data (parametric) and uncertainty over the return distribution, 2) propagate both through the Bellman equation, 3) track both with neural networks (i.e., high-capacity function approximators), and 4) use both to improve exploration.\footnote{As a side contribution, we introduce the Initial Return Entropy (IRE) as a measure of task exploration difficulty. See Appendix \ref{ire}.} 

The remainder of this paper is organized as follows. In Section 2 we provide a general introduction to Bayesian deep learning and distributional reinforcement learning. In Section 3, we discuss parametric and return uncertainty, and identify their potential for exploration. Section 4 discusses their implementations for policy evaluation with neural networks, and also discusses how to derive a policy from the learned distributions based on Thompson sampling. Sections 4, 5 and 6 show experimental results, discuss future work, and draw conclusions, respectively.

\section{Pre-liminaries}
\subsection{Bayesian deep learning} \label{bayesian}
Bayesian neural networks \citep{mackay2003information} represent the uncertainty in the model through posterior distributions over the model parameters. Assume we observe some random variables $X$ and $Y$ and are interested in the conditional distribution $p(Y|X)$. We introduced a neural network $p_\phi(Y|X)$ with parameters $\phi \in \Phi$ to estimate this conditional distribution. In the Bayesian setting, we treat the model parameters $\phi$ as random variables themselves. Given an observed dataset $\mathcal{H}$, we may use the posterior distribution over model parameters $p(\phi|\mathcal{H})$ to obtain the posterior predictive distribution 

\begin{equation}
p(y^\star|x^\star,\mathcal{H}) = \int p(y^\star|x^\star,\phi) p(\phi|\mathcal{H}) \mathrm{d}\phi \label{eq_postpred}
\end{equation}

for a new observed datapoint $x^\star$. In the non-linear neural networks of practical interest, the posterior distribution $p(\phi|\mathcal{H})$ is analytically intractable. \citet{gal2016dropout} showed that the well-known empirical procedure drop-out actually produces a Monte-Carlo approximation to Eq. \ref{eq_postpred}, providing samples from the posterior predictive distribution by simply retaining drop-out during test time (prediction). We use this technique in this paper, and discuss alternative methods for Bayesian inference in neural networks in the Future work section. 

\subsection{Distributional reinforcement learning} \label{distr_rl}
In reinforcement learning (RL) \citep{sutton1998reinforcement} agents are studied that interact with an unknown environment with the goal to optimize some long-term performance measure. The framework adopts a Markov Decision Process (MDP) given by the tuple $\{\mathcal{S},\mathcal{A},\mathcal{T},\mathcal{R},\gamma\}$. At every time-step $t$ we observe a state $s_t \in \mathcal{S}$ and pick an action $a_t \in \mathcal{A} = \{ 1...N_\mathcal{A} \} $, for $N_\mathcal{A}$ available discrete actions. The MDP follows the transition dynamics $s_{t+1} = \mathcal{T}(\cdot|s_t,a_t) \in \mathcal{S}$ and returns rewards $r_t = \mathcal{R}(s_t,a_t) \in \mathbb{R}$. For this work, we assume a discrete action space and deterministic transition and reward functions. 

We act in the MDP according to a stochastic policy $\pi$, i.e. $a \sim \pi(\cdot|s) \in \mathcal{P}(\mathcal{A})$. The discounted return $Z^\pi(s,a)$ from a state-action pair $(s,a)$ is a {\it random process} given by

\begin{equation} 
Z^\pi(s,a) = \sum_{t=0}^{\infty} \gamma^t r_t, \quad \quad  s_{t+1} = \mathcal{T}(\cdot|s_t,a_t),a_{t+1} \sim \pi(\cdot|s_{t+1}),s_0=s,a_0=a \label{eq_return}
\end{equation}

for discount factor $\gamma \in [0,1]$. We emphasize that the return $Z^\pi$ is a random variable, where the distribution of $Z^\pi$ is induced by the stochastic policy (as we assume a deterministic environment). We may rewrite equation \ref{eq_return} into a recursive form, known as the {\it distributional Bellman equation} (omitting the $\pi$ superscript from now on):

\begin{equation} 
Z(s,a) = r_t + \gamma Z(s',a'), \quad \quad s' = \mathcal{T}(\cdot|s,a),a' \sim \pi(\cdot|s'). \label{eq_distr_bellman}
\end{equation}

Note that the equality sign represents {\it distributional equality} here \citep{engel2005reinforcement}. We are now ready to define the action-value function. Denote by $\mathbb{E}_\pi$ the expectation over all traces induced by the policy $\pi$. Applying this operator to $Z(s,a)$ defines the state-action value $Q(s,a) = \mathbb{E}_\pi [Z(s,a)]$. Applying this operator to Eq. \ref{eq_return} gives

\begin{equation}
Q(s,a) = r_t + \gamma \mathbb{E}_{s' = \mathcal{T}(\cdot|s,a),a' \sim \pi(\cdot|s')} [Q(s',a') ] \label{eq_bellman}
\end{equation}

which is known as the Bellman equation \citep{sutton1998reinforcement}. Most RL papers actually start-off from Eq. \ref{eq_bellman}. We present the current introduction to emphasize that the mean action value $Q(s,a)$ is a quantity that we estimate by sampling from an underlying return distribution $p(Z|s,a)$.\footnote{We empirically observe that the shape of this return distribution strongly differs between domains. This matters because the shape of the return distribution also influences how easily we can estimate its expectation, or some other quantity like an upper confidence bound, from samples. For example, a long, thin right tail in the return distribution - as frequently the case in RL with only a few `good' traces - may give our mean estimate high variance (it would actually need importance sampling). In Appendix \ref{ire} we visualize return distributions for some well-known domains, and also introduce the {\it Initial Return Entropy} as a measure of task exploration difficulty.}

We approximate the action-value (distribution) with a deep neural network. We write $Q_\phi(s,a)$ for a network predicting a (point estimate) action-value, and $p_\phi(Z|s,a)$ for a network approximating the entire return distribution. To learn the state-action value RL algorithms follow variants of a scheme known as {\it generalized policy iteration} (GPI) \citep{sutton1998reinforcement}. GPI iterates between policy evaluation, in which we calculate a new estimates $\Psi(s,a)$ of the state-action value based on (new) sample data (e.g., for one-step SARSA $\Psi(s,a) = r(s,a) + \gamma Q(s',a')$), and policy improvement, in which we use the estimate $\Psi(s,a)$ to improve the policy (whether with a value-based, actor-critic or policy gradient algorithm). 

\section{Distributional perspective on exploration} \label{valuefunctions}
We will now argue for a probabilistic perspective on value functions and exploration. There are two distributions that might be useful from an exploration point of view: 1) the statistical parametric uncertainty of the mean action value, and 2) the distribution of the return.

\begin{figure}[t]
  \centering
      \includegraphics[width = 0.9\textwidth]{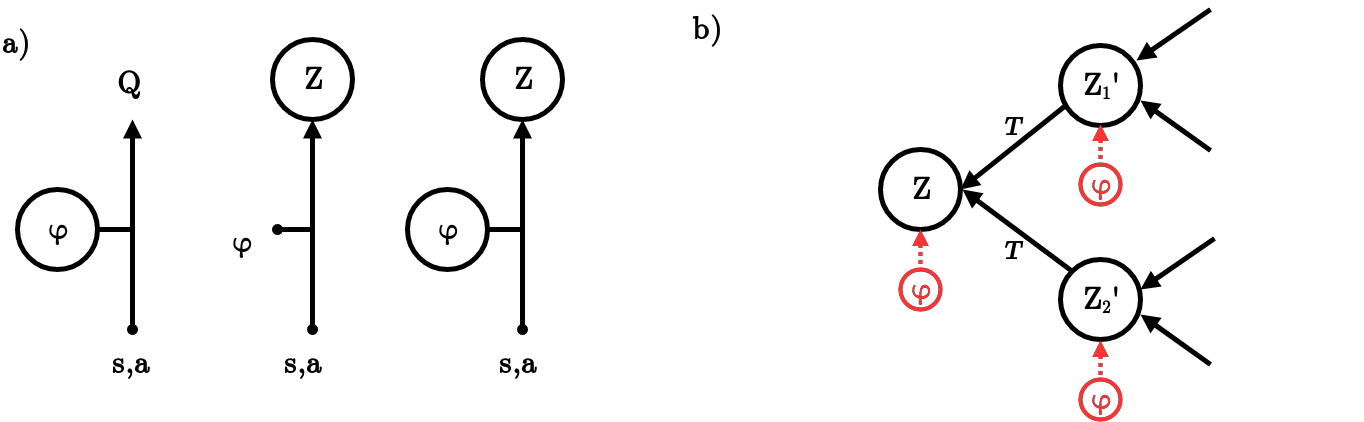}
  \caption{\small {\bf a)} Three types of neural networks with different uncertanties/probabilitiy distributions. Circles are probabilistic nodes. Left: parametric uncertainty over the mean action-value. Middle: propagating (return) distributions for point estimate parameters. Right: parametric uncertainty over propagating distribution (= Double Uncertain Value Network). {\bf b)} Illustration of propagating distributions. Subscripts identify unique state-action pairs. We initialize all state-action pairs with a prior parametric uncertainty $p(\phi)$ and prior output distribution $p_\phi(Z)$. Then, for a new observed transition, we want to update our estimates of $p_\phi(Z)$ at the current state action pair by propagating the distribution of the next node $p(Z')$ through the Bellman operator $T$ (instead of just propagating the mean). For this work, we consider two quantities to propagate: i) the return distribution at the next node $p_\phi(Z')$ (for point estimate $\phi$), or ii) the parametric uncertain return distribution at the next node $p(Z') = \int p_\phi(Z') p(\phi|\mathcal{H}) \mathrm{d} \phi$. Arrows point backwards because we focus on the direction of uncertainty propagation/back-up (which runs in the different direction than our exploration). }
    \label{fig_duvn}
\end{figure}

\paragraph{Parametric uncertainty of the mean} Given a policy the state-action value $Q(s,a)$ is a scalar number by definition, as it is an expectation over all possible future traces. However, from a statistical point of view it makes sense to treat our {\it estimate} of $Q(s,a)$ as a random variable, as we need to approximate it from a finite number of samples. We call this the parametric uncertainty. \newline
Parametric exploration, i.e. acting optimistic with respect to the uncertainty of the mean action-value, has been very successful in the bandit setting. However, it has been sparsely applied to RL (see Appendix \ref{relatedwork} for related work). We believe this is due to a fundamental complication regarding uncertainties in RL, which has only been identified by \citet{dearden1998bayesian} before. Bandits are one-step decision problems with pay-offs originating from a stationary distribution, which makes the value approximation an ordinary supervised learning problem. However, in RL the target distribution is highly non-stationary. A standard target like $r + \gamma Q(s',a')$ falsely assumes that $Q(s',a')$ is known, while it is actually uncertain itself. Therefore, repeatedly visiting a state-action pair should not makes us certain about its value if we are still uncertain about what to do next. In other words: {\it the state-action value certainty depends on the future policy certainty}. Standard parametric uncertainty cannot account for this problem (the `local' parametric uncertainty will converge as if it is supervised learning), and we somehow need to propagate the uncertainty of future state-action pairs' value (the `global' uncertainty) back through the Bellman equation. An illustration is seen in Fig. \ref{fig_duvn}b right, where the uncertainty in $\phi$ influences both the current and future value estimates (ignoring the $Z$ distributions in that graph for now, as the need to propagate the parametric uncertainty $p(\phi)$ over timesteps already applies when we learn mean action-values).

\paragraph{Return distribution} Standard RL, and also the parametric uncertainty introduced above, usually deal with the {\it mean} action-value $Q(s,a)$. However, from an exploration point of view, it makes more sense to learn the full {\it return distribution} $p(Z|s,a)$. Note that we still focus on deterministic environments. Therefore, the distribution over returns is solely induced by our own policy. As we may modify our own policy, it makes sense to act optimistically with respect to the return distribution we observe. As an illustration, consider a state-action pair with particular mean action value estimate $Q(s,a)$. It really matters whether this average originates from a highly varying return, or from consistently the same return. It matters because our policy may {\it influence} the shape of this distribution, i.e. for the highly varying returns we may actively transform the distribution towards the good returns. In other words, what we really care about in deterministic domains is the best return, or the upper end of the return distribution.\footnote{For stochastic domains the return distribution has additional noise for which we do want to act on the expectation.}

It turns out that both challenges focus around propagating either parametric uncertainty and/or return distributions through the Bellman equation (Fig \ref{fig_duvn}b). The overall idea is to memorize the propagating {\it global} MDP uncertainties in a neural network, which makes them {\it locally} available at action selection time. We thereby avoid the need for any forward planning (to get global information), and our approach is entirely model-free. 

\section{Double Uncertain Value Networks} \label{seq_duvn}

\subsection{Policy evaluation}
We now discuss three probabilistic policy evaluation approaches that incorporate the uncertainties introduced in the previous section: 1) (local) parametric uncertainty only, 2) return distribution only, and 3) both combined. The respective network structures are illustrated in Fig. \ref{fig_duvn}a. Implementation details are provided in Appendix \ref{appendix_implementation}.

\paragraph{Parametric uncertainty only} To estimate our parametric uncertainty we may use any type of Bayesian inference method suitable for neural networks. For this paper we consider the Bayesian dropout \citep{gal2016improving}, as it has a very simple practical implementation (see Sec. \ref{bayesian}). This gives us a sample from the posterior predictive distribution of the mean action value: $p(Q|s,a,\mathcal{H}) = \int Q_\phi(s,a) p(\phi|\mathcal{H}) \mathrm{d}\phi$. The associated network structure is visualized in Figure \ref{fig_duvn}a, left. 

\paragraph{Return distribution only}
We next consider the problem of learning return distributions instead of mean action-values. For this work we will assume that the return distribution $p(Z|s,a)$ can be approximated by a Gaussian. Therefore, we modify our neural network to output the distribution parameters $\mu^Z(s,a)$ and $\sigma^Z(s,a)$, where clearly $\mu^Z(s,a) = Q(s,a)$. Note that the network parameters $\phi$ are point estimates now. The associated network structure is visualized in Figure \ref{fig_duvn}a, middle. \newline
During policy evaluation we need to estimate distributional targets instead of point estimate targets. We will construct bootstrap estimators based on the distributional Bellman equation (Eq. \ref{eq_distr_bellman}). The derivation for the mean $\mu^Z(s,a) = Q(s,a)$ is well-known from standard RL, so we focus on propagating the return standard deviation through the distributional Bellman equation:

\begin{align}
\mathrm{Sd} \Big[ p(Z|s,a) \Big] &= \mathrm{Sd}\Big[ r(s,a) + \gamma \sum_{a' \in \mathcal{A}} \pi(a'|s') p(Z|s',a')\Big] = \gamma \sum_{a' \in \mathcal{A}} \pi(a'|s') \mathrm{Sd}\Big[p(Z|s',a')\Big]
\end{align}

because $\gamma \geq 0$, $\pi(a|s) \geq 0$, and we assume the next state distributions are independent so we may ignore the covariances.\footnote{For random variables $X,Y$ and scalar constants $a,b,c$ we have: $\mathrm{Var}[a + b X + c Y] = b^2 \hspace{0.1cm} \mathrm{Var}[X] + c^2 \hspace{0.1cm} \mathrm{Var}[Y] + 2bc \hspace{0.1cm} \mathrm{Cov}[X,Y]$.} We see that the standard deviation of $p(Z|s,a)$ is a linear combination of the standard deviations $p(Z|s',a')$ (one timestep ahead), reweighted by the policy probabilities and shrunken by $\gamma$. We approximate the sum over the policy probabilities $\pi(a'|s')$ by sampling from our policy (as is the usual solution in RL, which will be right in expectation over multiple traces). The network may then be trained to move the current predictions closer to these targets, for example with a squared loss

\begin{equation}
L(\phi) = \Big(r(s,a) + \gamma \mu_\phi^Z(s',a') - \mu_\phi^Z(s,a)\Big)^2 + \Big(\gamma \sigma_\phi^Z(s',a') - \sigma_\phi^Z(s,a)\Big)^2 \label{eq_loss}
\end{equation}

where we as usual fix the bootstrap predictions at $(s',a')$, i.e. the training gradients w.r.t. $\phi$ are blocked there. This approach can be seen as a form of analytic approximate return propagation with a (heuristic) distributional loss (see Appendix \ref{related_return} for other distributional losses). Similar ideas with approximate return propagation were recently explored with discrete network output distributions \citep{bellemare2017distributional}, which may also accommodate for propagating multimodality.

A second, more simple propagation method which we also experimented with is sampling-based propagation. In that setting we sample $M$ values  $z_m(s',a') \sim p_\phi(Z|s',a')$, push these through the Bellman operator to construct $\Psi_m(s,a) = r(s,a) + \gamma z_m(s',a')$, and train our network on this collection of samples with, e.g., a maximum likelihood loss. This may require more samples and be less accurate, but it will also work for complicated network output distributions (like deep generative models) for which analytic propagation and projection is infeasible. Results of this approach are not shown, but were comparable to the results with approximate return propagation shown in Section \ref{results}.

\paragraph{Parametric uncertainty over return distributions}
We finish with the observation that both ideas may actually naturally be combined in one function approximator (Fig \ref{fig_duvn}a, right). Note that we can now propagate both the return distribution and its parametric uncertainty at the next timestep, i.e. we are effectively propagating {\it uncertain return distributions} (the parametric uncertainty over the network output distribution). Starting from a sampled transition now, we want to propagate the return distribution weighted over the parametric uncertainty at the next timestep:

\begin{equation}
Z(s,a) = \int \Big[r + \gamma Z_\phi(s',a') \Big] p(\phi|\mathcal{H}) \mathrm{d} \phi = r + \gamma \int Z_\phi(s',a') p(\phi|\mathcal{H}) \mathrm{d} \phi
\end{equation}

Besides that, the same distributional Bellman propagating machinery applies as above.\footnote{Sample from $p(\phi|\mathcal{H})$ at the next time-step, make network predictions $\mu_\phi^Z(s',a')$ and $\sigma_\phi^Z(s',a')$, and do the Bellman propagation. Repeated sampling of $\phi$ does Monte Carlo integration over $p(\phi|\mathcal{H})$, as a numerical integration like in \citet{dearden1998bayesian} is infeasible for the neural network setting.} We refer to the general mechanism of uncertainty propagation (parametric, return or both) as {\it Bellman uncertainty}. \newline
The appearance of the network, with both uncertainty over the network parameters $\phi$ and over the output distribution to track the propagating (uncertain) return distributions, makes us refer to it as the Double Uncertain Value Network (DUVN) (Fig \ref{fig_duvn}a, right). The intuition is that during early learning we will mostly be propagating uncertainty, while with converging distributions we will eventually start propagating true return distributions.

In summary, we identified three types of probabilistic policy evaluation algorithms (with the three associated network structures visualized in Fig. \ref{fig_duvn}a):\footnote{We could think of another algorithm that does propagate (i.e., has a probabilistic network output), but only propagates the parametric uncertainty of the mean at the next time-step $p(Q'|\mathcal{H}) = \int Q'_\phi p(\phi|\mathcal{H}) \mathrm{d}\phi$ (and not the entire return distribution). We did not come up with such an algorithm, but concurrently with our work, \citet{o2017uncertainty} did focus on this problem. See Related Work.}
\begin{enumerate}
\item The (local) parametric uncertainty of the mean value: $p(Q|s,a,\mathcal{H}) = \int Q_\phi(s,a) p(\phi|\mathcal{H}) \mathrm{d}\phi$.
\item The (propagating) distribution of the return: $p_\phi(Z|s,a,\mathcal{H})$ (with point parameters $\phi$).
\item Both, (propagating) uncertain return distr.: $p(Z|s,a,\mathcal{H}) = \int p_\phi(Z|s,a,\mathcal{H}) p(\phi|\mathcal{H}) \mathrm{d}\phi$.
\end{enumerate}

\subsection{Policy improvement}
We now describe how to use any of these distributions to naturally balance exploration versus exploitation, based on Thompson sampling \citep{thompson1933likelihood} (see Appendix \ref{illustration} as well). To generalize notation, we introduce a new random variable $\Theta$ with distribution $p(\Theta|s,a)$ to capture any of the three policy evaluation distributions introduced in the previous section. We write $p(\boldsymbol{\Theta}|s) = \prod_{a^\star \in \mathcal{A}} p(\Theta|s,a^\star)$ for the joint action-value distribution in a state $s$, where we assume the posterior distributions per action are independent. Thompson sampling selects action $a$ with probability equal to:

\begin{equation}
\pi(a|s) = \int p(\Theta_{a}>\Theta_{a^\star \ne a}) p(\boldsymbol{\Theta}|s) \mathrm{d} \boldsymbol{\Theta}
\end{equation} 
 
where $\Theta_a = \Theta(s,a)$ and $\Theta_{a^\star \ne a}$ notational convention for $\Theta(s,a^\star) \forall a^\star \in \mathcal{A},a^\star \ne a$. In words, we choose action $a$ with probability equal to the probability that the specific action is the optimal one when averaging over all uncertainty in the joint distribution $p(\boldsymbol{\Theta}|s)$. The practical implementation of Thompson sampling is very simple, as we may just sample from $p(\Theta|s,a)$ for every $a$ and argmax over these values:

\begin{enumerate}
\item Sample $\phi \sim p(\phi)$ (or equivalently a dropout mask). 
\item Sample $Z(s,a^\star) \sim p_{\phi}(Z|s,a^\star) \quad \forall \quad a^\star \in \mathcal{A}$.
\item Select $a = \argmax_{a^\star \in \mathcal{A}} Z(s,a^\star)$.
\end{enumerate}

If we do not consider parametric uncertainty, then we ignore the first sampling step and just use the current parameter point estimates. If we do not consider the Bellman uncertainty, then we replace the second sampling step with a deterministic prediction $Q_\phi(s,a)$.

Thompson sampling is not the only possible choice to make decisions under uncertainty, but it has shown good empirical performance in the bandit literature \citep{chapelle2011empirical}. It naturally performs policy improvement, as it gradually starts to prefer the better actions when the distributions start narrowing/converging. We thereby hope to improve on the instability of greedy policy improvement (see also \citet{bellemare2017distributional}) or undirected exploration. Ideally, the uncertain return distribution would gradually narrow and for a deterministic environment eventually converge to a Dirac distribution on the optimal value function.

\begin{figure}[t]
  \centering
      \includegraphics[width = 0.82\textwidth]{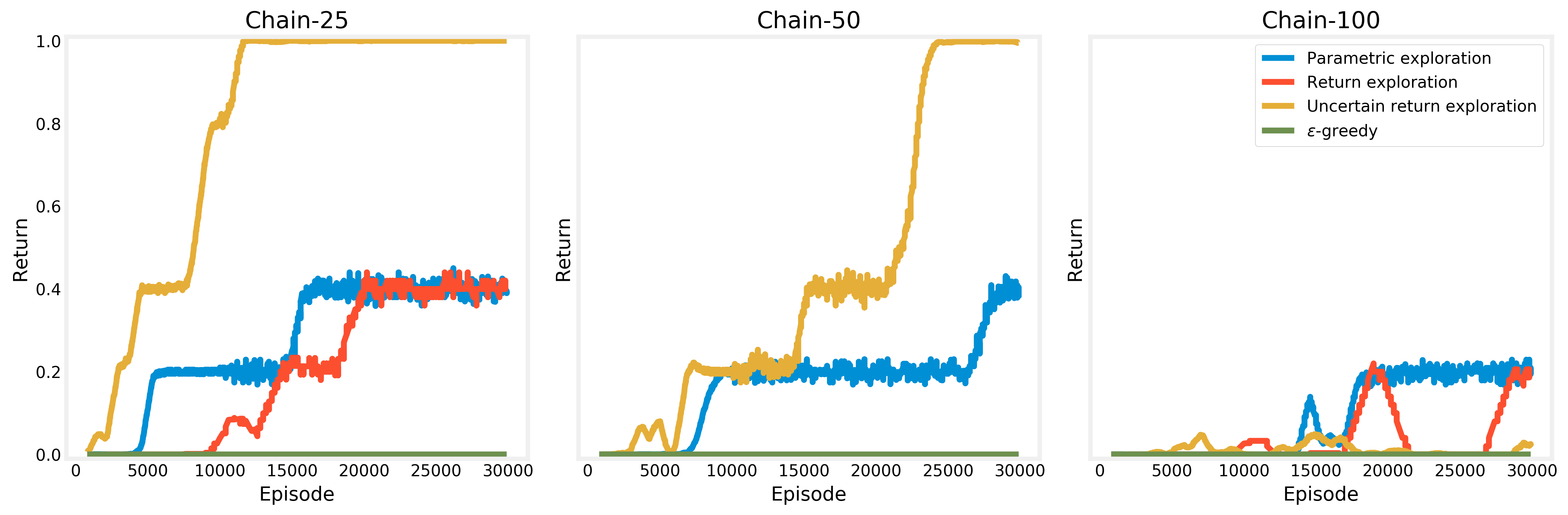}
  \caption{\small Learning curves on Chain domain for Thompson sampling on parametric uncertainty, return distribution and uncertain return distribution versus $\epsilon$-greedy exploration ($\epsilon = 0.05$). Plots progress row-wise for increased depth of the Chain, i.e. increased exploration difficulty. Note that the correct action at each state in the chain is initialized at random (i.e. not always action 2, as in the visualization in Fig. \ref{chainfigure}). Results averaged over 5 repetitions.}
    \label{fig_results_chain}
\end{figure}

\section{Experiments} \label{results}
We now evaluate the different types of probabilistic policy evaluation in combination with Thompson sampling exploration. We refer to Thompson sampling on the three types of discussed policy evaluation as {\it parametric exploration}, {\it return exploration}, and {\it uncertain return exploration}. Experimental details are provided in Appendix \ref{appendix_implementation}.

We first consider the Chain domain (Appendix \ref{chain}, Figure \ref{chainfigure}). The domain consists of a chain of states of length $N$, with two available actions at each state. The only trace giving a positive, non-zero reward is to select the `correct' action at every step. The correct action per step is determined at domain initialization by sampling from a uniform Bernoulli. The domain has a strong exploration challenge, which grows exponentially with the length of the chain (see Appendix \ref{chain}).

Learning curves for the Chain domain are shown in Fig. \ref{fig_results_chain}, for different lengths of the chain. First of all, we note that the $\epsilon$-greedy strategy does not learn in this domain at all (not even for the short length). The three probabilistic approaches do explore, with best performance for the uncertain return exploration. In the longest chain, of length 100, all probabilistic exploration methods also get trouble solving the domain. However, they do see some rewards, which makes us hypothesize this could be an issue of stabilization (more than that the exploration does not work). See Appendix \ref{extra_results} for results when the correct action is always the same, as in the original variants of this problem \citep{osband2016deep}.

\begin{figure}[t]
  \centering
      \includegraphics[width = 0.82\textwidth]{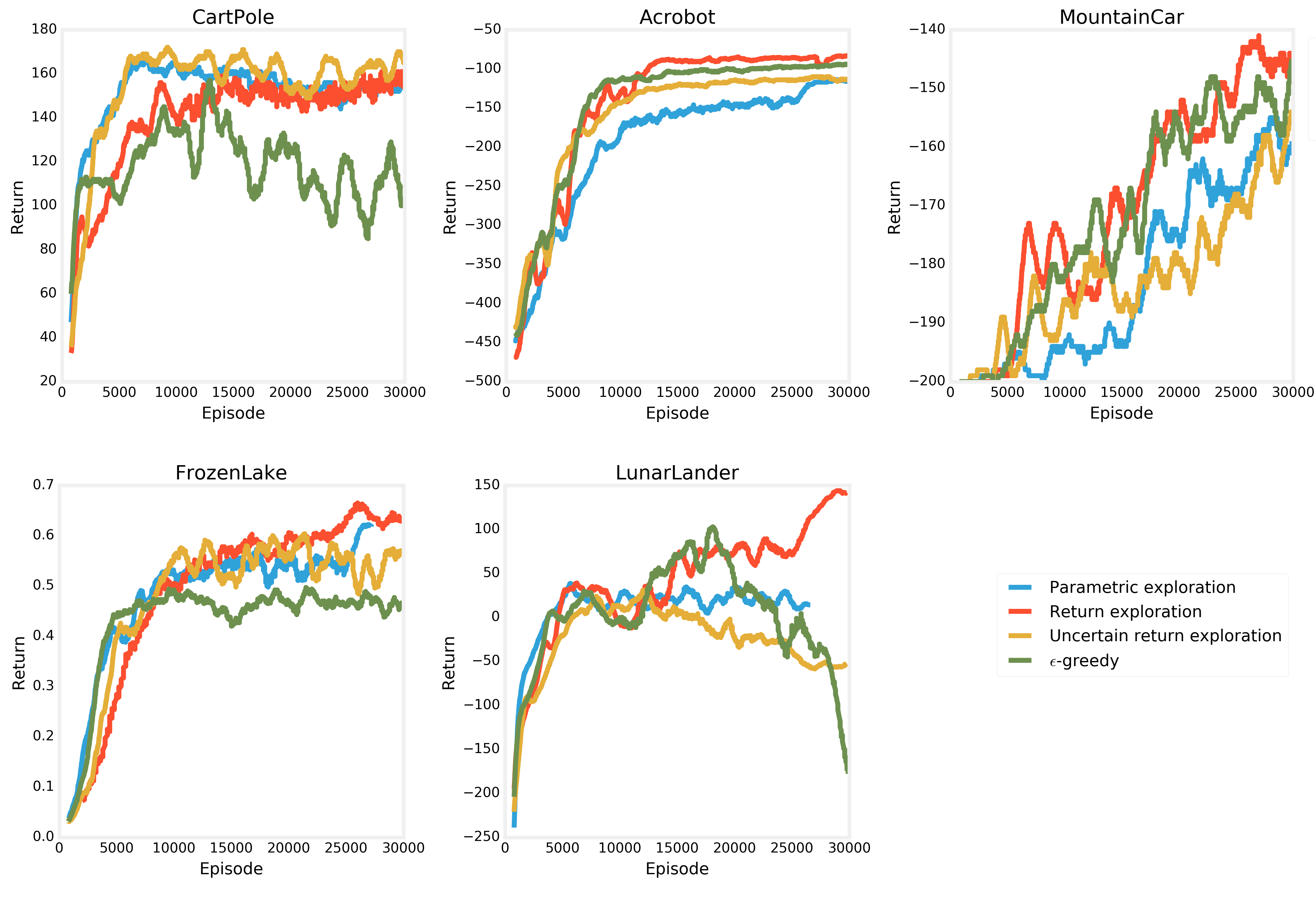}
  \caption{\small Learning curves for parametric exploration, return exploration and uncertain return exploration on different OpenAI Gym environments. Results averaged over 5 repetitions.}
    \label{fig_results_other}
\end{figure}

We next test our method on a set of tasks from the OpenAI Gym repository (Fig. \ref{fig_results_other}). We see that our exploration methods manage to learn on all domains. The achieved end policies all reflect good policies on each problem. $\epsilon$-greedy exploration is a bit unstable on some domains (CartPole, LunarLander), but generally performs reasonable as well. We note that the uncertainty exploration methods, which have a completely different exploration mechanism compared to $\epsilon$-greedy exploration, never really perform worse on these domains.

We hypothesize these domains have too much structure and are not challenging enough to show the same exploration difference as seen for the Chain domain. Future work should address more challenging (high-dimensional) exploration problems. We also want to stress that probabilistic exploration will not always outperform undirected methods, especially not on domains with relatively simple exploration. Uncertainty methods will generally create a cautious agent, that first wants to properly verify all parts of the domain. In contrast, undirected exploration agents may exploit sooner, which can be beneficial in domains with non-deep exploration (i.e., with quick rewards).

\section{Future work} \label{future}
We identify several directions for future work:
\begin{enumerate}
\item {\bf Other types of Bayesian inference in neural networks} (for parametric uncertainty): we hypothesize that the Bayesian drop-out may be too unstable and tedious to tune, as we sometimes observed in our experiments as well. Potentially, different methods to approximate the posterior over the network parameters (e.g., \citet{welling2011bayesian}) may improve estimation of parametric uncertainty.  
\item {\bf More expressive output distributions} (for Bellman uncertainty propagation): for this work we only experimented with Gaussian distributions for propagation. Recently, \citet{bellemare2017distributional} studied return distribution propagation with categorical distributions, which more naturally accommodate for multi-modality (see Related Work as well). Another extension could involve more expressive continuous network distributions, e.g. based on conditional variational inference \citep{moerland2017learning}. 
\item {\bf Continuous action-spaces}: the current implementations only focussed on discrete action spaces, where Thompson sampling can easily be applied by maintaining a distribution per action and enumerating all actions for action selection. Extension to the continuous setting would require either directly propagating policy uncertainty, or learning a parametric policy whose distribution mimics the uncertainty in the value function.
\item {\bf Stochastic environments}: this paper entirely focussed on domains with deterministic reward and transition functions, which makes the return distribution only induced by the stochastic policy. In stochastic domains the return distribution will have additional noise for which we do want to act on the expectation, to prevent continuing to act optimistically with respect to something we can't influence.
\end{enumerate}

Finally, we want to stress that the RL algorithms in this paper are entirely model-free. The uncertainty theme also appears in model-based RL, where it is useful/necessary to track the uncertainty on an estimated transition and/or reward function \citep{deisenroth2011pilco,depeweg2017uncertainty}. This parametric {\it model} uncertainty is different from the parametric {\it value/policy} uncertainty studied in this work, but our ideas may be extended to the model-based setting as well (which would add another source of uncertainty).  

\section{Conclusion}
This paper introduced Double Uncertain Value Networks (DUVN), which, to the best of our knowledge, is the first algorithm that 1) distinguishes between uncertainty due to limited data (parametric) and uncertainty over the return distribution, 2) propagates both through the Bellman equation, 3) tracks both with neural networks (i.e., high-capacity function approximators), and 4) uses both to improve exploration. We implemented the DUVN algorithm with Bayesian dropout for the parametric uncertainty and a Gaussian distribution for the Bellman uncertainty propagation. The main appeal of this implementation is its simplicity: any deep Q-network implementation can be easily extended as in this work by adding drop-out to the neural network layers and specifying a Gaussian output distribution instead of a mean-only prediction. This should take no more then a few lines of code in most automatic differentiation software packages. We showed that, even for the vanilla implementation, we at least match or improve undirected exploration performance on a variety of problems, and drastically improve performance on an exploration heavy domain (Chain). We believe further improvements in the distributional approach to RL, e.g. with more expressive network output distributions that capture multi-modality, is a promising direction for RL exploration research. 

\clearpage
{\footnotesize
\bibliographystyle{apalike}
\bibliography{uncertainty}

\begin{thebibliography}{}

\bibitem[Auer et~al., 2002]{auer2002finite}
Auer, P., Cesa-Bianchi, N., and Fischer, P. (2002).
\newblock {Finite-time analysis of the multiarmed bandit problem}.
\newblock {\em Machine learning}, 47(2-3):235--256.

\bibitem[Bellemare et~al., 2016]{bellemare2016unifying}
Bellemare, M., Srinivasan, S., Ostrovski, G., Schaul, T., Saxton, D., and
  Munos, R. (2016).
\newblock {Unifying count-based exploration and intrinsic motivation}.
\newblock In {\em {Advances in Neural Information Processing Systems}}, pages
  1471--1479.

\bibitem[Bellemare et~al., 2017]{bellemare2017distributional}
Bellemare, M.~G., Dabney, W., and Munos, R. (2017).
\newblock {A distributional perspective on reinforcement learning}.
\newblock {\em arXiv preprint arXiv:1707.06887}.

\bibitem[Blundell et~al., 2015]{blundell2015weight}
Blundell, C., Cornebise, J., Kavukcuoglu, K., and Wierstra, D. (2015).
\newblock {Weight uncertainty in neural networks}.
\newblock {\em arXiv preprint arXiv:1505.05424}.

\bibitem[Brafman and Tennenholtz, 2002]{brafman2002r}
Brafman, R.~I. and Tennenholtz, M. (2002).
\newblock {R-max-a general polynomial time algorithm for near-optimal
  reinforcement learning}.
\newblock {\em Journal of Machine Learning Research}, 3(Oct):213--231.

\bibitem[Chapelle and Li, 2011]{chapelle2011empirical}
Chapelle, O. and Li, L. (2011).
\newblock {An empirical evaluation of thompson sampling}.
\newblock In {\em {Advances in neural information processing systems}}, pages
  2249--2257.

\bibitem[Dearden et~al., 1999]{dearden1999model}
Dearden, R., Friedman, N., and Andre, D. (1999).
\newblock {Model based Bayesian exploration}.
\newblock In {\em {Proceedings of the Fifteenth conference on Uncertainty in
  artificial intelligence}}, pages 150--159. Morgan Kaufmann Publishers Inc.

\bibitem[Dearden et~al., 1998]{dearden1998bayesian}
Dearden, R., Friedman, N., and Russell, S. (1998).
\newblock {Bayesian Q-learning}.
\newblock In {\em {AAAI/IAAI}}, pages 761--768.

\bibitem[Deisenroth and Rasmussen, 2011]{deisenroth2011pilco}
Deisenroth, M. and Rasmussen, C.~E. (2011).
\newblock {PILCO: A model-based and data-efficient approach to policy search}.
\newblock In {\em {Proceedings of the 28th International Conference on machine
  learning (ICML-11)}}, pages 465--472.

\bibitem[Depeweg et~al., 2017]{depeweg2017uncertainty}
Depeweg, S., Hern{\'a}ndez-Lobato, J.~M., Doshi-Velez, F., and Udluft, S.
  (2017).
\newblock {Uncertainty Decomposition in Bayesian Neural Networks with Latent
  Variables}.
\newblock {\em arXiv preprint arXiv:1706.08495}.

\bibitem[Engel et~al., 2003]{engel2003bayes}
Engel, Y., Mannor, S., and Meir, R. (2003).
\newblock {Bayes meets Bellman: The Gaussian process approach to temporal
  difference learning}.
\newblock In {\em {Proceedings of the 20th International Conference on Machine
  Learning (ICML-03)}}, pages 154--161.

\bibitem[Engel et~al., 2005]{engel2005reinforcement}
Engel, Y., Mannor, S., and Meir, R. (2005).
\newblock {Reinforcement learning with Gaussian processes}.
\newblock In {\em {Proceedings of the 22nd international conference on Machine
  learning}}, pages 201--208. ACM.

\bibitem[Gal, 2016]{gal2016uncertainty}
Gal, Y. (2016).
\newblock {\em {Uncertainty in deep learning}}.
\newblock PhD thesis, PhD thesis, University of Cambridge.

\bibitem[Gal and Ghahramani, 2016]{gal2016dropout}
Gal, Y. and Ghahramani, Z. (2016).
\newblock {Dropout as a Bayesian approximation: Representing model uncertainty
  in deep learning}.
\newblock In {\em {international conference on machine learning}}, pages
  1050--1059.

\bibitem[Gal et~al., 2016]{gal2016improving}
Gal, Y., McAllister, R.~T., and Rasmussen, C.~E. (2016).
\newblock {Improving PILCO with bayesian neural network dynamics models}.
\newblock In {\em {Data-Efficient Machine Learning workshop}}, volume 951, page
  2016.

\bibitem[Ghavamzadeh and Engel, 2007a]{ghavamzadeh2007a}
Ghavamzadeh, M. and Engel, Y. (2007a).
\newblock {Bayesian actor-critic algorithms}.
\newblock In {\em {Proceedings of the 24th international conference on Machine
  learning}}, pages 297--304. ACM.

\bibitem[Ghavamzadeh and Engel, 2007b]{ghavamzadeh2007b}
Ghavamzadeh, M. and Engel, Y. (2007b).
\newblock {Bayesian policy gradient algorithms}.
\newblock In {\em {Advances in neural information processing systems}}, pages
  457--464.

\bibitem[Guez et~al., 2012]{guez2012efficient}
Guez, A., Silver, D., and Dayan, P. (2012).
\newblock {Efficient Bayes-adaptive reinforcement learning using sample-based
  search}.
\newblock In {\em {Advances in Neural Information Processing Systems}}, pages
  1025--1033.

\bibitem[Houthooft et~al., 2016]{houthooft2016vime}
Houthooft, R., Chen, X., Duan, Y., Schulman, J., {De Turck}, F., and Abbeel, P.
  (2016).
\newblock {Vime: Variational information maximizing exploration}.
\newblock In {\em {Advances in Neural Information Processing Systems}}, pages
  1109--1117.

\bibitem[Kearns and Singh, 2002]{kearns2002near}
Kearns, M. and Singh, S. (2002).
\newblock {Near-optimal reinforcement learning in polynomial time}.
\newblock {\em Machine Learning}, 49(2-3):209--232.

\bibitem[Kocsis and Szepesv{\'a}ri, 2006]{kocsis2006bandit}
Kocsis, L. and Szepesv{\'a}ri, C. (2006).
\newblock {Bandit based monte-carlo planning}.
\newblock In {\em {ECML}}, volume~6, pages 282--293. Springer.

\bibitem[MacKay, 2003]{mackay2003information}
MacKay, D.~J. (2003).
\newblock {\em {Information theory, inference and learning algorithms}}.
\newblock Cambridge university press.

\bibitem[Mannor et~al., 2004]{mannor2004bias}
Mannor, S., Simester, D., Sun, P., and Tsitsiklis, J.~N. (2004).
\newblock {Bias and variance in value function estimation}.
\newblock In {\em {Proceedings of the twenty-first international conference on
  Machine learning}}, page~72. ACM.

\bibitem[Mannor et~al., 2007]{mannor2007bias}
Mannor, S., Simester, D., Sun, P., and Tsitsiklis, J.~N. (2007).
\newblock {Bias and variance approximation in value function estimates}.
\newblock {\em Management Science}, 53(2):308--322.

\bibitem[Mannor and Tsitsiklis, 2011]{mannor2011mean}
Mannor, S. and Tsitsiklis, J. (2011).
\newblock {Mean-variance optimization in Markov decision processes}.
\newblock {\em arXiv preprint arXiv:1104.5601}.

\bibitem[Matthias et~al., 2017]{Plappert17Parameter}
Matthias, P., Rein, H., Prafulla, D., Szymon, S., {Richard Y.}, C., Xi, C.,
  Tamim, A., Pieter, A., and Marcin, A. (2017).
\newblock {Parameter Space Noise for Exploration}.
\newblock {\em arXiv preprint arXiv:1706.01905}.

\bibitem[Moerland et~al., 2017]{moerland2017learning}
Moerland, T.~M., Broekens, J., and Jonker, C.~M. (2017).
\newblock {Learning Multimodal Transition Dynamics for Model-Based
  Reinforcement Learning}.
\newblock {\em arXiv preprint arXiv:1705.00470}.

\bibitem[Morimura et~al., 2012]{morimura2012parametric}
Morimura, T., Sugiyama, M., Kashima, H., Hachiya, H., and Tanaka, T. (2012).
\newblock {Parametric return density estimation for reinforcement learning}.
\newblock {\em arXiv preprint arXiv:1203.3497}.

\bibitem[O'Donoghue et~al., 2017]{o2017uncertainty}
O'Donoghue, B., Osband, I., Munos, R., and Mnih, V. (2017).
\newblock {The Uncertainty Bellman Equation and Exploration}.
\newblock {\em arXiv preprint arXiv:1709.05380}.

\bibitem[Osband et~al., 2016]{osband2016deep}
Osband, I., Blundell, C., Pritzel, A., and {Van Roy}, B. (2016).
\newblock {Deep exploration via bootstrapped DQN}.
\newblock In {\em {Advances in Neural Information Processing Systems}}, pages
  4026--4034.

\bibitem[Osband et~al., 2014]{osband2014generalization}
Osband, I., {Van Roy}, B., and Wen, Z. (2014).
\newblock {Generalization and exploration via randomized value functions}.
\newblock {\em arXiv preprint arXiv:1402.0635}.

\bibitem[Rasmussen et~al., 2003]{rasmussen2003gaussian}
Rasmussen, C.~E., Kuss, M., et~al. (2003).
\newblock {Gaussian Processes in Reinforcement Learning.}
\newblock In {\em {NIPS}}, volume~4, page~1.

\bibitem[Sobel, 1982]{sobel1982variance}
Sobel, M.~J. (1982).
\newblock {The variance of discounted Markov decision processes}.
\newblock {\em Journal of Applied Probability}, 19(4):794--802.

\bibitem[Stadie et~al., 2015]{stadie2015incentivizing}
Stadie, B.~C., Levine, S., and Abbeel, P. (2015).
\newblock {Incentivizing exploration in reinforcement learning with deep
  predictive models}.
\newblock {\em arXiv preprint arXiv:1507.00814}.

\bibitem[Sutton and Barto, 1998]{sutton1998reinforcement}
Sutton, R.~S. and Barto, A.~G. (1998).
\newblock {\em {Reinforcement learning: An introduction}}, volume~1.
\newblock MIT press Cambridge.

\bibitem[Tamar et~al., 2016]{tamar2016learning}
Tamar, A., {Di Castro}, D., and Mannor, S. (2016).
\newblock {Learning the variance of the reward-to-go}.
\newblock {\em Journal of Machine Learning Research}, 17(13):1--36.

\bibitem[Thompson, 1933]{thompson1933likelihood}
Thompson, W.~R. (1933).
\newblock {On the likelihood that one unknown probability exceeds another in
  view of the evidence of two samples}.
\newblock {\em Biometrika}, 25(3/4):285--294.

\bibitem[van Hoof et~al., 2017]{van2017generalized}
van Hoof, H., Tanneberg, D., and Peters, J. (2017).
\newblock {Generalized exploration in policy search}.
\newblock {\em Machine Learning}, 106(9-10):1705--1724.

\bibitem[Welling and Teh, 2011]{welling2011bayesian}
Welling, M. and Teh, Y.~W. (2011).
\newblock {Bayesian learning via stochastic gradient Langevin dynamics}.
\newblock In {\em {Proceedings of the 28th International Conference on Machine
  Learning (ICML-11)}}, pages 681--688.

\bibitem[White, 1988]{white1988mean}
White, D. (1988).
\newblock {Mean, variance, and probabilistic criteria in finite Markov decision
  processes: a review}.
\newblock {\em Journal of Optimization Theory and Applications}, 56(1):1--29.

\end{thebibliography}
}

\clearpage

\clearpage
\appendix

\clearpage
\section{Related work} \label{relatedwork}
Exploration is a widely studied topic in reinforcement learning. We will discuss work based on parametric (value/policy) uncertainty, return distributions/uncertainty, and add some context on other exploration approaches (count-based/intrinsic motivation) and other uncertainty methods in RL (uncertainty in model-based RL).

\subsection{Parametric uncertainty of the mean} \label{related_parametric}
This research direction uses the uncertainty of the mean action value, either from a frequentist or Bayesian direction, to direct exploration. Exploration is usually based on 'optimism under uncertainty'. Parametric uncertainty has been extensively studied in the {\it bandit} setting, which are environments with a single state, multiple actions and unknown, stochastic rewards. Succesful approaches are UCB \cite{auer2002finite}, which acts on the upper confidence bound of a frequentist confidence interval, and Thompson sampling \cite{thompson1933likelihood}, which is also studied in this paper. 

There are a few extensions of these ideas to the RL/MDP setting. The first occurrence of parametric uncertainty in RL seems to be Bayesian Q-learning by \citet{dearden1998bayesian}. The authors use tabular Q-learning with normal distributions and conjugate updating. They are also the only ones that explicitly identify the necessity to propagate parametric uncertainty from future states. Their exploration is based on either Thompson sampling (which they call Q-value sampling), while they also consider myopic value of perfect information (VPI) as another exploration strategy. 

\citet{osband2014generalization} extended these ideas to the linear function approximation setting with randomized least-squares value iteration (RLSVI). In the neural network context, parametric uncertainty based on variational inference was studied for bandits by \citet{blundell2015weight}. \citet{gal2016dropout} studied the use of dropout uncertainty for parametric value uncertainty similar to our work, but did not consider any propagation, nor the distribution over returns. \citet{osband2016deep} also studied parametric exploration in RL with neural networks, using the non-parametric bootstrap (i.e., a frequentist approach to uncertainty estimation, not to be confused with the use of the term `bootstrapping' in RL). 

Concurrently with the present paper, \citet{o2017uncertainty} also identified the need to propagate parametric uncertainty over timesteps. Their approach is based on a variance estimate, which has a similar role as the $\sigma$ in our Gaussian uncertainty propagation. Their neural network implementation derives the local parametric uncertainty estimates from the linearity of their last network layer and frequentist uncertainty estimates known from linear regression. This contrasts to our Bayesian approach to parametric uncertainty. Moreover, they still propagate uncertainty about the mean action value, and do not consider the returns as in our paper. 

There is more work that does track uncertainty for policy evaluation, but does not use these for policy improvement / exploration. Most of these have focussed on Gaussian Process regression \citep{engel2003bayes}. \citet{rasmussen2003gaussian} uses two Gaussian Processes: one to track the parametric uncertainty in the value, and a second one to model the uncertainty in the transition model. However, the paper still uses a greedy policy improvement. The Gaussian Process approach was also extended to continuous action spaces. There are actor-critic \citep{ghavamzadeh2007a} and policy search \citep{ghavamzadeh2007b} algorithms that track the uncertainty in the {\it gradients}, but again only to stabalize the update, not to direct exploration.

The idea of exploration based on parametric uncertainty also connects to the difference between {\it action space} and {\it parameter space} / episode-based exploration \citep{Plappert17Parameter,van2017generalized}. Most (undirected) exploration methods, like $\epsilon$-greedy and Boltzmann, inject exploration noise at the action space level. However, it can be beneficial to inject the noise at parameter level instead, usually because it allows you to retain a particular noise setting over multiple steps (e.g. an entire episode). The risk of action-space exploration noise is that the agent has to redecide at every timestep and therefore cannot stick with an exploration decision. The effect might be jittering behaviour between exploration and exploitation steps. This has also been identified as the challenge of ensuring `deep' exploration \cite{osband2016deep}. We have not considered this problem in this paper, but it could for example be implemented by fixing the dropout mask over an entire episode.

We also want to note that the exact same exploration problem occurs in classical (tree) search. Succesful Monte Carlo Tree Search (MCTS) algorithms, like Upper Confidence Bounds for Trees (UCT) \cite{kocsis2006bandit}, act on the upper confidence bound of a frequentist confidence interval of the value at each state-action pair. The overlap between reinforcement learning and (model-based) search has been identified for long \cite{sutton1998reinforcement}, where RL i) does not assume an a-priori known environment model, and ii) usually includes a parametric function approximator to represent the value/policy, while search stores these in a tree structure (which is technically an effective sparse form of tabular representation). But besides that, the same exploration themes appear in both fields.

\subsection{Return uncertainty} \label{related_return}
While the distributional Bellman equation (Eq. \ref{eq_distr_bellman}) is certainly not new \cite{sobel1982variance,white1988mean}, nearly all RL research has focussed on the mean action-value. Most papers that do study the underlying return distribution study the 'variance of the return'. \citet{engel2005reinforcement} learned the distribution of the return with Gaussian Processes, but did not use it for exploration. \citet{tamar2016learning} studied the variance of the return with linear function approximation. \citet{mannor2011mean} theoretically studies policies that bound the variance of the return. 

The variance of the return does not need to be used with 'optimism under uncertainty', and actually has more frequently been considered for {\it risk-sensitive RL}. In several scenarios we may want to avoid incidental large negative pay-offs, which can e.g. be desastrous for a real-world robot, or in a financial portfolio. \citet{morimura2012parametric} studied parametric return distribution propagation as well. They do risk-sensitive exploration by softmax exploration over {\it quantile} Q-functions (also known as the {\it Value-at-Risk} (VaR) in financial management literature). Their distribution losses are based on KL-divergences (including Normal, Laplace and skewed Laplace distributions), which could be a better distributional loss than the heuristic loss in Eq. \ref{eq_loss}. However, their implementations do remain in the tabular setting.

Recently, \citet{bellemare2017distributional} theoretically studied the distributional Bellman operator. The authors show that the operator is still a contraction in the policy evaluation setting, but not a contraction in any distribution metric for the control setting. They hypothesize this is due to the `inherent instability of greedy updates' in the Bellman optimality operator. Their algorithm (called C51) uses a categorical distribution to propagate returns distributions, which may more easily accommodate for multimodality compared to the Gaussian distribution used in this work. C51 backs-up the complete Bellman distributions, but they do not use these for exploration. Their methods nevertheless improves over all other previous deep Q-networks on Atari games.

\subsection{Count-based Exploration \& Intrinsic Motivation} Count-based exploration uses a slightly different incentive for exploration, focussing or rewarding regions of state-space that have not been visited (often). These ideas were extensively studied in the {\it tabula rasa} setting, e.g. R-max \cite{brafman2002r} and Explicit-Explore or Exploit ($E^3$) \cite{kearns2002near}. \citet{guez2012efficient} explicitly plans ahead using Monte Carlo Tree Search over uncertain transition dynamics models. Applications in high-dimensional domains include \cite{stadie2015incentivizing} and  \cite{bellemare2016unifying}.

Intrinsic motivation generalizes this notion of novelty to any internal reward for {\it domain-independent} characteristics, i.e. next to the domain-dependent external reward function. An example is rewarding actions that decreases the parametric uncertainty in the transition model \citep{houthooft2016vime}. Alltogether, this class of exploration methods usually depends on the ability to learn good transition models (from limited data), a problem which is not trivial itself \citep{deisenroth2011pilco,depeweg2017uncertainty,moerland2017learning}.

A theoretical problem with count-based / intrinsic motivation approaches can be that they change the RL objective itself. For example, bonuses on novelty might make an agent continue to visit a region of state-space where the value functions are already very certain, yet not all states are frequently visited yet (like continuously walking around a room to view it from all angles, which gives a new visual state each time). Nevertheless, intrinsic motivation-based approaches hold the state-of-the-art on challenging exploration problems like Montezuma's Revenge \citep{bellemare2016unifying}.

\subsection{Uncertainty in model-based RL}
All work in this paper only considered model-free RL. However, similar issues with uncertainty appear when learning the transition and/or reward function, known as `model-based RL'. \citet{dearden1999model} was again the first to address this problem for the tabular setting. \citet{mannor2004bias,mannor2007bias} studied two environment sources of variance that may influence the return distribution: the `internal variance' due to a stochastic environment, and the `parametric (model) variance' due to bias in the environment model. Neither of these were considered in this work, but both may be added. These ideas were studied in neural networks by \citet{depeweg2017uncertainty}, who instead uses the terms {\it empistemic uncertainty} for the model bias and {\it aleatoric uncertainty} for the inherent environment noise/stochasticity. Their approach, which infers distributions on neural network parameters to capture model bias, and uses expressive output distributions to capture true environment stochasticity, actually has a similar structure as our Double Uncertain Value Network (in some sense, they learn a `Double Uncertain Transition Network'). Of course, transition model learning does not involve any uncertainty propagation (it is a well-defined supervised learning problem). Finally, \citet{gal2016improving} used Bayesian drop-out, as considered in this work for parametric value uncertainty, to track parametric model uncertainty. 

\clearpage
\section{Initial Return Entropy (IRE) as a measure of initial exploration difficulty} \label{ire}
Define the initial return distribution (IRD) $p^\text{init}(Z)$ as the distribution over trace returns $Z \in \mathbb{R}$ when sampling an initial state $s^\text{init}$ from some initial state distribution and following a uniform random policy from there on. For undirected exploration, the uniform policy is the best policy we can specify until we start encountering varying returns. Define the initial return entropy (IRE) as the entropy of this distribution: 

\begin{equation}
H(Z) = \int p^\text{init}(Z) \cdot \log \Big( p^\text{init}(Z) \Big) \mathrm{d} Z
\end{equation}

\begin{figure}[h]
  \centering
      \includegraphics[width = 0.9\textwidth, height=0.62\textwidth]{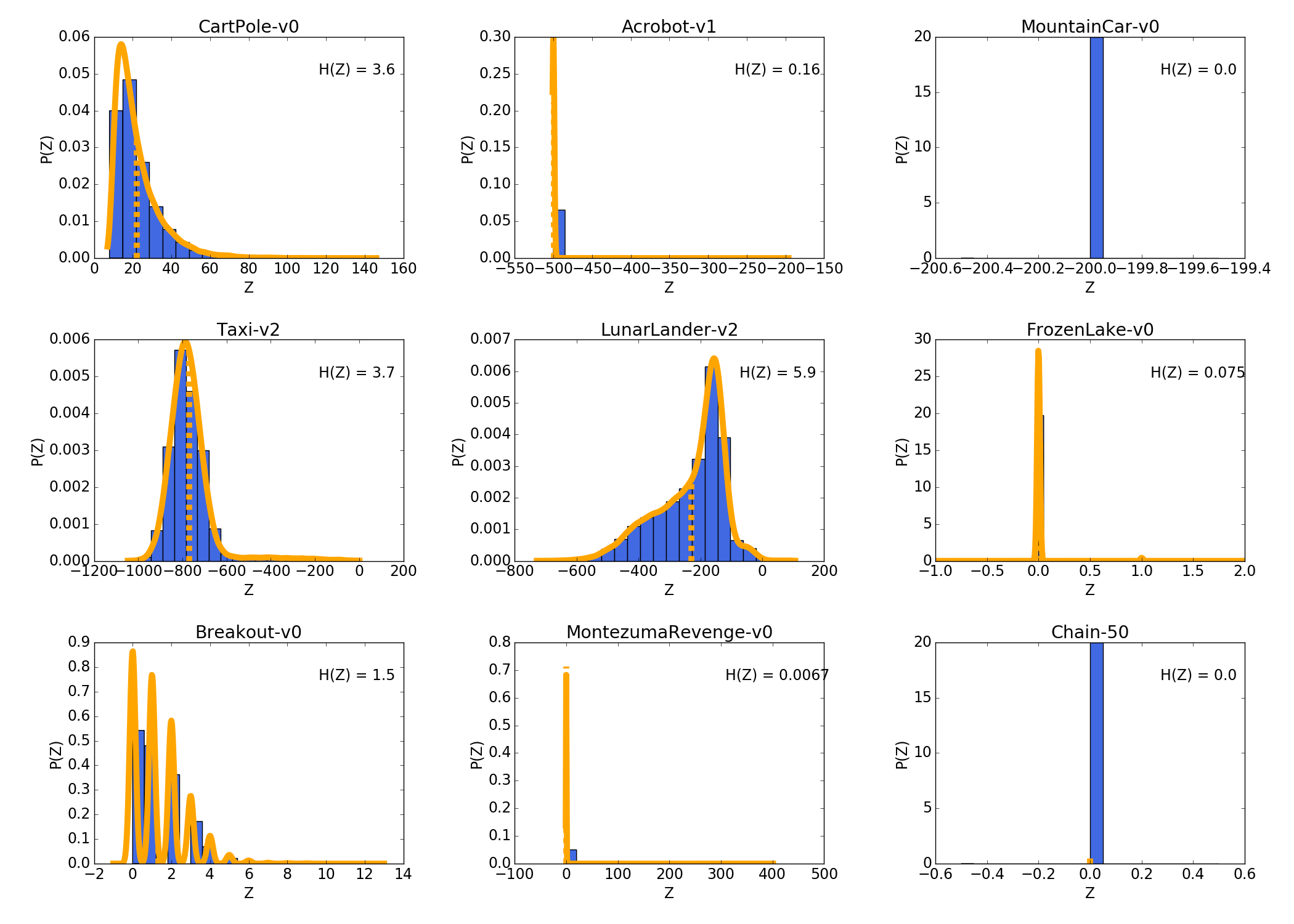}
  \caption{\small Return distributions from the initial state in different environments for a uniform random policy. Histogram produced over 50.000 traces of maximum 500 steps. The first 8 domains are directly taken from the OpenAI Gym. The Chain domain is introduced in Appendix \ref{chain}. Orange line is a kernel density estimate, with the vertical dashed line its empirical mean (a Monte Carlo estimate of $Q(s,a)$ under a uniform random policy. The top-right display the initial return entropy (IRE) estimate for the domain.}
    \label{returndistribution}
\end{figure}

We propose that the IRE is an interesting measure of the domain exploration difficulty, where lower values indicate a higher exploration challenge. Figure \ref{returndistribution} shows the IRD and IRE for various domains from the OpenAI Gym repository. We see quite large differences between the shape of these distributions. Importantly, some domains with hard exploration, for example the Atari game Montezuma's Revenge, show a very spiked IRD and therefore low IRE. The challenge of such domains is that {\it nearly all initial traces give the same return}, which makes it hard to ever find a first indication of where to go. In many of such domains nearly all traces then give 0 reward, but for example Mountain Car shows all traces giving -200 reward (there is a -1 penalty per timestep, and Gym caps MountainCar episodes at length 200 by default). The entropy of the return distribution is of course robust against such reward function translations, making it a stable measure of initial exploration difficulty.

We do not propose this is the only measure of domain exploration difficulty. For example, a well-known exploration challenge is choosing between a small suboptimal pay-off and exploring further to obtain a potential higher reward. This type of exploration challenge is not accurately reflected in the IRE, as simple early rewards spread out the initial return distribution and may falsely suggest the domain is easy. There are of course many more dimensions that influence the RL task difficulty, like the state and action space cardinality, but the IRE nicely illustrates why a low-dimensional task like the Chain (Appendix \ref{chain}) can actually be quite challenging.  

\clearpage

\section{Illustration of undirected versus directed exploration} \label{illustration}
\begin{figure}[h]
  \centering
      \includegraphics[width = 0.9\textwidth]{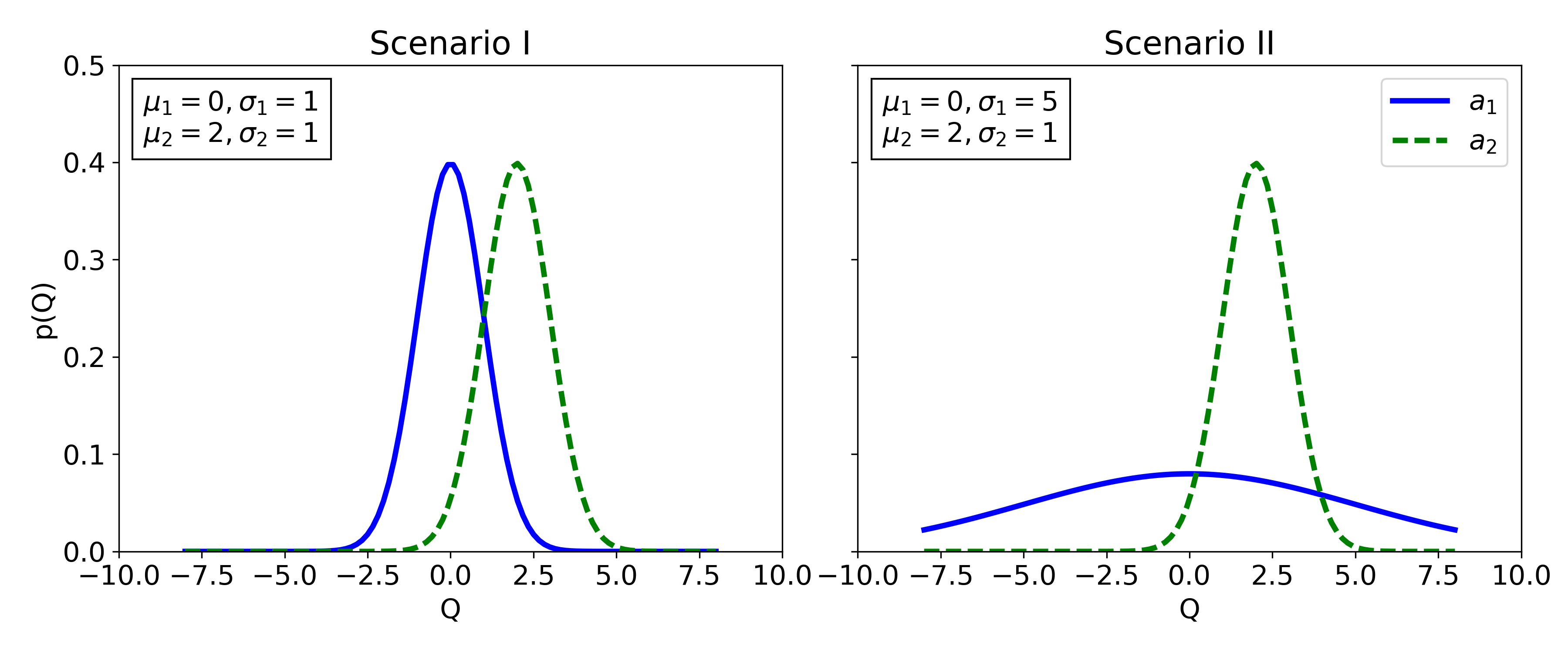}
  \caption{\small Example posterior value distributions for two available actions. Scenario I (left): Action 1 (blue solid line) has $\mu_1=0$, $\sigma_1=1$, Action 2 (green dashed line) has $\mu_2=2$, $\sigma_2=1$. Scenario II (right): The same except for $\sigma_1=5$.}
    \label{appendix_example}
\end{figure}

We will quickly elaborate on the difference between undirected exploration methods, like $\epsilon$-greedy and Boltzmann exploration, and directed methods, like Thompson sampling, in a theoretical example. Consider two available actions which, given some observed data $\mathcal{H}$, both have some posterior action-value distribution $p(Q|\mathcal{H})$. Figure \ref{appendix_example} shows two scenario's, I (left) and II (right). The only difference between both scenario's is our uncertainty about the value of action $a_1$: in the second scenario we are much more uncertain about its true value. We now compare how $\epsilon$-greedy, Boltzmann and Thompson sampling will act in both scenario's. The main point will be that undirected methods cannot leverage the uncertainty information. 

{\bf 1. $\boldsymbol{\epsilon}$-greedy} exploration only uses the distribution means and will act the same in both scenarios, preferring action 2 and selecting action 1 with (small) probability $\epsilon$. 

{\bf 2. Boltzmann} (soft-max) exploration is usually seen as more subtle, gradually preferring actions with a higher pay-off. Boltzmann does consider the numerical scale of the action means, including their difference (something $\epsilon$-greedy ignores):

$$ \pi_{Boltzmann}(a|s) = \frac{e^{\mu_a}}{\sum_{a^\star \in \mathcal{A}} e^{\mu_{a^\star}}} $$

However, it still acts the same in scenario I and II, because Boltzmann approximates a distribution over both actions by still {\it only considering their means}. Although the softmax returns a probability distribution over actions, this should can {\bf not} be interpreted as their uncertainty.\footnote{A similar phenomenon happens with the softmax and cross-entropy loss in classification tasks. The outputs of this softmax are also frequently falsely interpreted as a measure of uncertainty over classes \citep{gal2016uncertainty}. However, when we extrapolate (far) away from our observed data, one of the classes usually gets a high probability. It thereby appears as we are very certain, but since we have not observed any data in this region of input space, we should actually be very uncertain. This illustrates how point estimates over a discrete set {\it cannot} be transformed to uncertainties (what we need is an entire uncertainty/distribution per class output).} Another problem with Boltzmann action selection is that it is non-robust against translation of the reward function, making it tedious to tune. Therefore, many undirected implementations still prefer $\epsilon$-greedy exploration.

{\bf 3. Thompson sampling}, a directed exploration method, uses $\pi(a_1|s) = p(Q_1>Q_2)$. For the example with normal random variables $Q_1 \sim \mathcal{N}(\cdot|\mu_1,\sigma_1)$ and $Q_2 \sim \mathcal{N}(\cdot|\mu_2,\sigma_2)$, we can analytically calculate $P(Q_1>Q_2) = P(Q_1-Q_2>0)$. Define $X=Q_1-Q_2$, then $X$ will still have a normal distribution, and by standard laws of probability, $\mathbb{E}[X] = \mu_1 - \mu_2$, and $\text{Sd}[X] = \sqrt{(\sigma_1)^2 + (\sigma_2)^2}$. Applying this to the example, gives us for Scenario I (left) $\pi(a_1) = \mathcal{N}(Q_1 - Q_2>0|\mu_X=-1,\sigma_X=\sqrt{2}) \approx 0.08$, and for Scenario II (right) $\pi(a_1) = \mathcal{N}(Q_1 - Q_2>0|\mu_X=-1,\sigma_X = \sqrt{26}) \approx 0.35$. Note how Thompson sampling naturally assigns extra probability mass to action $a_1$ in Scenario II, where we are much more uncertain about its potential value. 

\clearpage
\section{Illustration of exploration challenge: Chain domain} \label{chain}
\begin{figure}[ht]
  \centering
      \includegraphics[width = 0.6\textwidth]{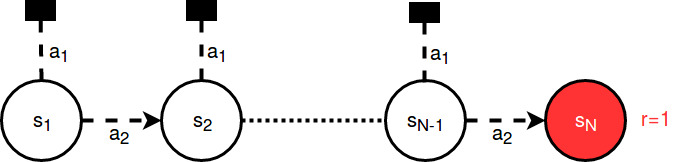}
  \caption{\small Chain domain. Example MDP where undirected exploration is highly inefficient. Based on \citep{osband2014generalization}.}
    \label{chainfigure}
\end{figure}

We now study the Chain domain, an example MDP (Fig. \ref{chainfigure}) that illustrates the difficulty of exploration with sparse rewards. This domain is also empirically studied in the results section of this paper. The MDP consists of a chain of states $\mathcal{S} = \{1,2...,N\}$. At each time step the agent has two available actions: $a_1$ (`left') and $a_2$ (`right'). At every step, one of both actions is the `correct' one, which deterministically moves the agent one step further in the chain. The wrong action terminates the episode. All states have zero reward except the final chain state $N$, which has $r=1$. Variants of these problem have been studied more frequently in RL \citep{osband2014generalization}. In the `ordered' implementation, the correct action is always the same (e.g. $a_2$), and the optimal policy is to always walk right. This is the variant illustrated in Fig. \ref{chainfigure} as well.

\citet{osband2014generalization} studied the expected regret for a variant of this scenario. We here present a different illustration, where we show the expected time until the first visit to the terminal state (i.e. the first non-zero trace in this domain).

\paragraph{Example 1.} {\it Let $l$ denote the number of episodes before we first reach state $N$. Clearly, before we  reach $N$ for the first time, we have seen no reward information, and undirected exploration will follow a uniform random policy. The probability of a trace reaching state $N$ under the uniform policy is $p = 2^{-(N-1)}$. Therefore, the number of episodes until we first reach $N$ follows a negative binomial distribution with success probability $p$, i.e. $l \sim NB(1,p)$. It follows that $\mathbb{E}[l] = \frac{1-p}{p} = 2^{(N-1)} - 1 $.}

Example 1 shows that, for undirected exploration on point estimates, the required number of exploratory episodes scales {\it exponentially} with the exploration depth $N$. Although this is clearly a simplified domain, it is important to note that this setting is actually very representative of the exploration problem in sparse reward domains. This is well visible in Fig. \ref{returndistribution}, where we can see the Chain domain having similar initial return distributions as for example Montezuma's Revenge, a game notorious for its challenging exploration.

\subsection{Additional results for ordered Chain} \label{extra_results}
The experiments section in the paper discusses the unordered Chain, where the correct action at every step is randomized. We here compare to the `ordered' Chain, where the correct action at every step is always $a_2$. Although this is the standard implementation in literature, we believe there is a systematic bias in this domain that makes them not really exponentially challenging for exploration. The problem is that the optimal policy has a network predicting $a_2$ at every step. This will happen too easily in a function approximator (like a neural network) due to its natural tendency to generalize. 

\begin{figure}[h]
  \centering
      \includegraphics[width = 1.0\textwidth]{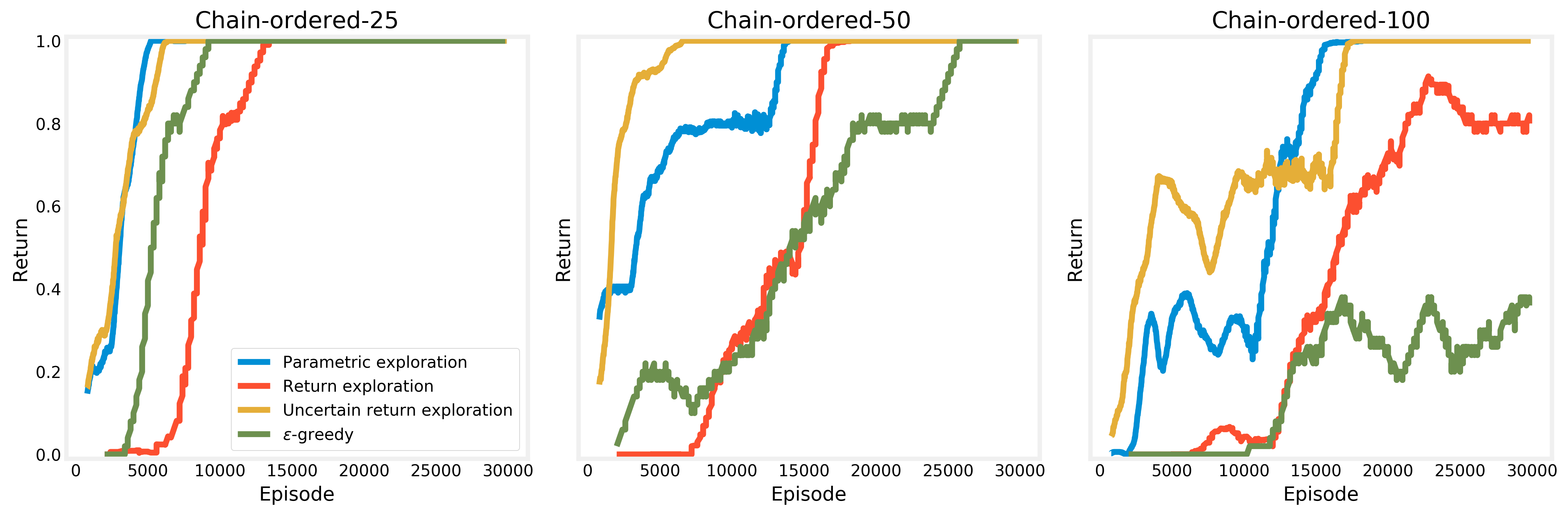}
  \caption{\small Learning curves on the ordered Chain domain.}
    \label{fig_results_orderedchain}
\end{figure}

We show the results on the ordered Chain in Fig. \ref{fig_results_orderedchain}. First of all, compared to the results in Fig. \ref{fig_results_chain}, we see that exploration is indeed much easier in the ordered problem. For example, $\epsilon$-greedy now also solves the problem, something which we would not expect from the exponential exploration time discussed above. Nevertheless, we see that the probabilistic exploration methods still outperform $\epsilon$-greedy, especially when the length of the chain increases. 

\clearpage

\section{Implementation Details} \label{appendix_implementation}
Network architecture consists of a 3 layer network {\it for each discrete action} with 128 nodes in each hidden layer and ReLu activations. For parametric uncertainty, each hidden layer has drop-out applied to its output, with $p_{keep}$ probability to keep a node. We use separate subnetworks per action to explicitly separate their uncertainty. For larger problems, the initial representation layers may be shared. Learning rates are fixed at 0.001 on all experiments. Optimization is performed with stochastic gradient descent using Adam updates in Tensorflow. For the experiments with parametric exploration (parametric uncertainty only) we train on a standard squared loss between new target and predicted mean action value, i.e. the first half of Eq. \ref{eq_loss}. We use a target network and replay database, where we replay $10\%$ of times in a prioritized way (by maintaining a separate prioritized replay queue based on the total temporal difference error in the previous time this trace was trained on). All domains (except for the Chain) are taken from the OpenAI Gym repository available at \url{https://github.com/openai/gym}.

All $\epsilon$-greedy experiments have $\epsilon$ fixed at 0.05 throughout learning. Drop-out rates were either $p_{keep}=0.75$ (for parametric uncertainty only) or $p_{keep}=0.90$ (for uncertain returns). All experiments used one-step SARSA (ie., on-policy updates with eligibility traces parameter $\lambda=0$ \citep{sutton1998reinforcement}), except for the MountainCar experiments, which use $\lambda=0.9$. Note that the ideas about eligibility traces and cutting traces equally apply to the propagation of distributions, i.e. they allow for quicker propagation over multiple timesteps (always at the risk of propagating on-policy, exploratory results too quickly/far).\footnote{We do believe that the uncertainty-based policies, like Thompson sampling, may also benefit work on cutting traces. Trace cutting is usually based on importance sampling ratios between the exploratory policy and the target policy. Thompson sampling may provide more realistic probabilities for exploratory actions, which may allow for more natural trace cutting. For example, $\epsilon$-greedy always strongly cuts a trace for every exploratory step, no matter whether the exploratory action is very close to the best one, or known to be very bad. In contrast, probabilistic policies will cut traces when other possible actions in the state have much uncertainty left, which should indeed stop the speed of our back-ups. A challenge is that our neural network implementation naturally samples a next action, but the associated probability of each action is not directly available. Of course, for a small discrete action space we could approximate it by repeatedly sampling from our policy.} \newline
Thompson sampling uses the same policy for exploration and evaluation. In some sense, proper uncertainty policies somewhat blur the line between on- and off-policy (where the behavioural/exploration policy differs from the target/evaluation policy), as there is just on reasonable probabilistic policy incorporating all uncertainty. Nevertheless, we could consider Thompson sampling exploration while evaluating with a policy that does act on some mean value again. 
\end{document}